\documentclass{article}

\PassOptionsToPackage{numbers, compress}{natbib}

\usepackage[preprint]{neurips_2023}

\usepackage[utf8]{inputenc} 
\usepackage[T1]{fontenc}    
\usepackage{hyperref}       
\usepackage{url}            
\usepackage{booktabs}       
\usepackage{amsfonts}       
\usepackage{nicefrac}       
\usepackage{microtype}      
\usepackage{xcolor}         
\usepackage{graphicx}

\usepackage{amsmath}
\usepackage{algorithm}
\usepackage{algpseudocode}

\title{\textsc{BatGPT}: A Bidirectional Autoregressive Talker from Generative Pre-trained Transformer}

\author{%
  Zuchao Li, Shitou Zhang, Hai Zhao\thanks{Corresponding author.}, Yifei Yang, Dongjie Yang \\
  School of Computer Science, Wuhan University\\
  Department of Computer Science and Engineering, Shanghai Jiao Tong University\\
  \texttt{zcli-charlie@whu.edu.cn, shitouzhang@whu.edu.cn, zhaohai@cs.sjtu.edu.cn} \\
}

\begin{document}

\maketitle

\begin{abstract}
    \textsc{BatGPT} is a large-scale language model designed and trained jointly by Wuhan University and Shanghai Jiao Tong University. It is capable of generating highly natural and fluent text in response to various types of input, including text prompts, images, and audio. In the modeling level, we employ a bidirectional autoregressive architecture that allows the model to efficiently capture the complex dependencies of natural language, making it highly effective in tasks such as language generation, dialog systems, and question answering. Moreover, the bidirectional autoregressive modeling not only operates from left to right but also from right to left, effectively reducing fixed memory effects and alleviating model hallucinations.
    In the training aspect, we utilize a parameter expansion strategy for leveraging the pre-training of existing models and employ reinforcement learning from both AI and human feedback, aimed at improving the model's alignment performance. Overall, these approaches significantly improve the effectiveness of \textsc{BatGPT}, and the model can be utilized for a wide range of natural language applications.
\end{abstract}

\section{Introduction}


In recent years, natural language processing has made remarkable progress, especially the large language model (LLM) pre-training, leading to the development of various language models capable of generating text with high fluency and naturalness. Language models are becoming increasingly impactful and crucial, as they underpin many applications integral to our daily lives. From generating human-like text for content creation to providing recommendations based on contextual understanding, language models serve as the backbone of numerous systems that improve the efficiency and effectiveness of workflows for millions of users.

The landscape of modern transformer-based language models has been continuously evolving. Among these models, the Generative Pre-trained Transformers (GPTs) \cite{Radford2018ImprovingLU, Radford2019LanguageMA, brown2020language} has emerged as one of the most prominent ones due to their ability to efficiently model complex patterns in natural language. GPT and its variants have brought about unprecedented capabilities in generating high-quality text, making strides in fluency, coherence, and generalization. 

Under the hood, the GPT-like models predominantly rely on learning representations with a causal language modeling objective. They are categorized as unidirectional models as they only utilize context from one direction. In contrast, bidirectional models \cite{devlin2018bert, liu2019roberta, raffel2020exploring} are trained with a denoising pre-training objective, such as masked language modeling, allowing them to utilize context from both directions for prediction.

To better align LLMs with different training objectives with a variety of downstream tasks, several key strategies have been implemented. T5 \cite{raffel2020exploring} reformulates different NLP tasks into text-to-text problems, thus facilitating a unified solution for diverse tasks. Further, instruction tuing models like FLAN \cite{wei2022finetuned} and T0 \cite{sanh2022multitask} leverages task-specific instructions during pre-training to enhance performance across various tasks. These instructions, integrated directly into the input, enable models to better align with task-specific objectives, leading to improved task completion and cross-task generalization. To close the gap between NLP tasks with human needs, reinforcement learning from human feedback (RLHF) \cite{bai2022training, ouyang2022training} has been employed. By providing models with feedback on their generated outputs, they can fine-tune their predictions to better align with human expectations. This iterative learning process enables the model to adapt its output based on the reward signal from the feedback, paving the way for more accurate and contextually relevant language generation. 

These advancements, in tandem, have contributed to the growing sophistication and capabilities of LLMs, pushing the boundaries of LLMs to unprecedented levels that are closer than ever to human-like performance. Recent iterations of these models have demonstrated exceptional proficiency across a wide range of language tasks, even surpassing human-level performance in several evaluations \cite{openai2023gpt4}.

Despite these advancements, current models still face limitations and challenges. One prominent issue is the "limited memory" effect, where the model's ability to maintain context dwindles with increasing sequence length. Additionally, these models often suffer from hallucinations, where they generate outputs not aligned with the input context, and face difficulties in capturing complex dependencies efficiently. These challenges impose barriers on the utilization of these models for certain high-stakes applications.

To overcome these limitations and build upon existing advancements, Wuhan University and Shanghai Jiao Tong University jointly developed \textsc{BatGPT}, a large-scale language model that utilizes a bidirectional autoregressive architecture for language generation, dialog systems, and question answering. In this paper, we present the design, modeling, and training of \textsc{BatGPT}, highlighting our bidirectional autoregressive architecture, parameter expansion strategy, and reinforcement learning approach for improving the model's alignment performance. Our results demonstrate that \textsc{BatGPT} is highly effective and versatile, making it a valuable alternative for various natural language applications.

\section{Related Work}

A diverse spectrum of pre-trained language models have been developed over the years, categorized mainly into autoregressive \cite{Radford2018ImprovingLU, Radford2019LanguageMA, brown2020language, lin2022fewshot}, autoencoding \cite{devlin2018bert, yang2020xlnet, liu2019roberta, clark2020electra, joshi2020spanbert}, and encoder-decoder models \cite{bi-etal-2020-palm, pmlr-v97-song19d, lewis-etal-2020-bart, pmlr-v119-zhang20ae, raffel2020exploring, sanh2022multitask}. These models, each with their unique design philosophy and strength, have collectively pushed forward the boundary of natural language understanding and generation.

T5 \cite{raffel2020exploring} was one of the early models that introduced the paradigm shift of formulating almost all NLP tasks as generation tasks. This approach was subsequently adopted and refined by instruction tuning models such as FLAN \cite{wei2022finetuned, chung2022scaling} and T0 \cite{sanh2022multitask}, which enhanced the performance across various tasks by enriching the pre-training phase with more task-specific instructions. Further, self-instruct \cite{wang2023selfinstruct} was proposed to address the limitations of human-written instructions in terms of quantity, diversity, and creativity by enabling models to generate their instructions via a bootstrapping framework. Recent models like Alpaca \cite{alpaca} and Guanaco \footnote{https://guanaco-model.github.io/}, which were fine-tuned from LLaMA \cite{touvron2023llama} using self-generated instructions, have shown promising performance, validating the effectiveness of self-instruct.

Insights into the scaling laws of LLMs have also provided valuable guidance for balancing the trade-off between training cost and model performance. \cite{Kaplan2020ScalingLF} showed that increasing model size, data size, and computational resources can lead to improved performance. This scaling law was further affirmed by models like GPT-3 \cite{brown2020language} and Gopher \cite{rae2022scaling}. While the Chinchilla study \cite{hoffmann2022training} showed that under the same computational budget, scaling down the model size yields uniformly and significantly better performance over models of larger sizes, indicating that most existing LLMs were far from adequately trained. This revised understanding of the scaling laws has guided the development of newer models such as BLOOM \cite{workshop2023bloom}, GLM-130B \cite{zeng2022glm130b}, and etc.

Another key effort to better align model capabilities with human needs is the integration of reinforcement learning from human feedback (RLHF) \cite{bai2022training, ouyang2022training}, which enables models to generate more helpful and less harmful output by learning from human feedback. Recent work \cite{bai2022constitutional} have further replaced human feedback with AI feedback as the reward signal, making it possible to control AI behavior with far fewer human labels.

\textsc{BatGPT} builds upon these advancements and aims at addressing some of the persistent limitations in the field. We introduce a bidirectional autoregressive architecture that not only enhances the model's capability in handling complex dependencies but also mitigates the limited memory issue. The parameter expansion method employed by \textsc{BatGPT} leverages the knowledge garnered during the pre-training of existing models, thus facilitating a significant reduction in time and computational costs. Inspired by RLHF models, \textsc{BatGPT}'s reinforcement learning approach further refines the alignment between model outputs and human expectations.

In summary, \textsc{BatGPT} addresses the limitations of its predecessors while incorporating their strengths. We hope the unique approach and advanced features of \textsc{BatGPT} can set a new benchmark in the field, not only presenting the potential to contribute to a wide range of natural language applications, but also paving the way for future model development.

\section{\textsc{BatGPT}}

\subsection{Bidirectional Autoregressive Pre-training}

\textsc{BatGPT} is pre-trained using a bidirectional autoregressive language modeling objective, a modification of the traditional autoregressive objective where the model learns to predict the next token based on all previously seen tokens in the sequence, from both the past and future directions. This makes the model capable of capturing dependencies in both the forward and backward context.

Let $x = (x_1, x_2, ..., x_T)$ denote a sequence of tokens of length $T$. The goal of \textsc{BatGPT} during pre-training is to maximize the joint probability of observing the entire sequence. More specifically, the model aims to predict each token $x_{t}$ given the preceding tokens $x_{<t}$ or the subsequent tokens $x_{>t}$.

Given the sequence $x$, the pre-trained \textsc{BatGPT}, denoted as $\pi_{\phi}^{\text{pretrain}}$, parameterized by $\phi$, outputs a distribution over possible tokens at each position from both ends of the sequence: $\pi_{\phi}^{\text{pretrain}}(x_t | x_{<t})$ and $\pi_{\phi}^{\text{pretrain}}(x_t | x_{>t})$. The objective for pre-training can be written as:

\begin{equation}
J^{\text{pretrain}}(\phi) = \mathbb{E}_{x \sim D^{\text{pretrain}}} \left[ \sum_{t=1}^{T} \log \pi_{\phi}^{\text{pretrain}}(x_t | x_{<t} ) + \sum_{t=1}^{T} \log \pi_{\phi}^{\text{pretrain}}(x_t | x_{>t}) \right],
\end{equation}

where $D^{\text{pretrain}}$ represents the distribution over the pre-training data and the expectation is taken over all sequences $x$ sampled from $D^{\text{pretrain}}$. By maximizing $J^{\text{pretrain}}$, \textsc{BatGPT} learns to capture the intricate linguistic patterns, semantics, and structure inherent in the vast array of data it is trained on, thus yielding more coherent and fluent outputs.

\subsection{Instruct Tuning}

Following the pre-training stage, \textsc{BatGPT} is further refined via instruction tuning. This process utilizes a wealth of prompted data in the form of $\mathrm{\langle prompt, response \rangle}$ pairs. These pairs serve as contextualized cues for the model to generate appropriate responses, thereby facilitating the alignment of \textsc{BatGPT}'s behavior with human instructions and expectations.

Specifically, the prompted dataset, denoted as $D^{\text{inst}}$, consists of pairs $(x, y)$, where $x$ stands for a prompt and $y$ for the corresponding response. The instruction tuning objective then becomes to optimize the following likelihood function:

\begin{equation}
J^{\text{inst}}(\phi) = \mathbb{E}_{(x, y) \sim D^{\text{inst}}} \left[ \log \pi_{\phi}^{\text{inst}}(y|x) \right],
\end{equation}

where $\pi_{\phi}^{\text{inst}}(y|x)$ represents the probability of generating the response $y$ given the prompt $x$ according to the instruction tuning updated model $\pi_{\phi}^{\text{inst}}$. In addition to this, \textsc{BatGPT} is further refined using multi-round dialogue data, which takes concatenated conversation history as input and last-round response as output. This focused training strategy is specifically devised to enhance \textsc{BatGPT}'s capability in comprehending long chats. It optimizes \textsc{BatGPT}'s ability to preserve conversational context over extended conversations, allowing for more coherent and in-depth dialogue generation.

The instruct tuning phase essentially tunes \textsc{BatGPT}'s parameters $\phi$ to better generate responses that align with the given prompts, enabling \textsc{BatGPT} to effectively process and appropriately respond to diverse instructions.

\subsection{Reinforcement Learning from Human Feedback}

Reinforcement Learning from Human Feedback (RLHF) forms a crucial component of \textsc{BatGPT} training pipeline. To make the aligning process more efficient and flexible, \textsc{BatGPT} learns not only from human feedback but also from feedback generated by other AI systems.

The underlying objective of RLHF is to optimize a reward model $R$ which is then used to train \textsc{BatGPT} through Proximal Policy Optimization (PPO) \cite{schulman2017proximal}. The reward model $R$ is trained on collected preference data, which consists of pairs of model-generated responses $(y, y')$ along with a preference $d \in \{-1, 0, 1\}$. Here, $d = -1$ denotes preference for $y$, $d = 1$ denotes preference for $y'$, and $d = 0$ represents equivalence. The collection of human preference data is detailed below. 

\paragraph{Preference Data Collection}

\begin{figure}
    \centering
    \includegraphics[width=1.0\linewidth]{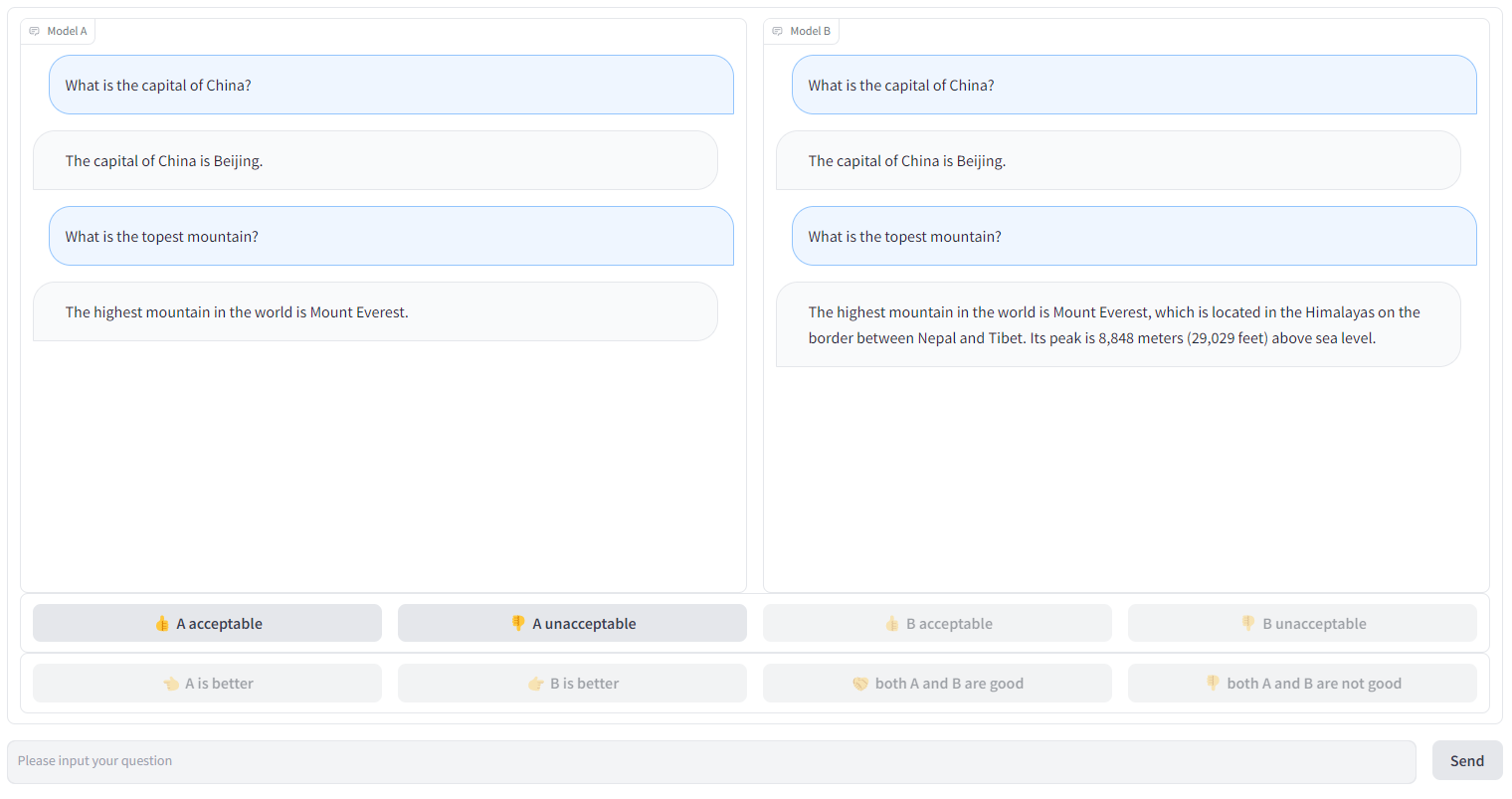}
    \caption{Annotation platform.}
    \label{fig:annotation}
\end{figure}

In order to train \textsc{BatGPT} via RLHF, it was necessary to gather an ample amount of preference data based on human judgement. To expedite this process, we developed a preference data annotation platform. The front-end page design presented to the annotators is shown in \ref{fig:annotation}. On this platform, the annotators are presented with two model outputs, denoted as $\mathcal{A}$ and $\mathcal{B}$, generated in response to the same instruction. These outputs can either be both from \textsc{BatGPT} or one from \textsc{BatGPT} and one from another LLM.

The platform provides the annotator with a predefined set of options to make the task of comparison more systematic and efficient. For evaluating the acceptability of the outputs, particularly focusing on potential harmfulness, the annotator is presented with four choices: "$\mathcal{A}$ is acceptable", "$\mathcal{A}$ is unacceptable", "$\mathcal{B}$ is acceptable", and "$\mathcal{B}$ is unacceptable".

In order to label the helpfulness of the outputs, the annotator is given another set of four options: "$\mathcal{A}$ is better", "$\mathcal{B}$ is better", "both $\mathcal{A}$ and $\mathcal{B}$ are good", and "both $\mathcal{A}$ and $\mathcal{B}$ are not good". The use of these predefined options streamlines the process of evaluating model outputs, reduces ambiguity, and increases the reliability of the feedback gathered, thus ensuring a more robust and effective training process for \textsc{BatGPT}.

Following the collection of human feedback, \textsc{BatGPT} employs AI systems to amass additional preference data. This is achieved through meticulously designed prompt templates where the feedback options align with what is presented to human annotators. The feedback gathered from both humans and AI, consolidated, creates an extensive preference dataset, which further enhances the depth and diversity of the training pool.

\paragraph{RLHF Training}

The reward model $R$ parameterized by $\theta$, predicts rewards $r(y)$ and $r(y')$ for each $(y, y', d)$ triplet in the preference dataset $D^{\text{pref}}$, where $r(y) = R(x, y)$ and $r(y') = R(x, y')$. The reward model is trained by minimizing the the following loss:

\begin{equation}
L^{\mathrm{R}}(\theta) = \mathbb{E}_{(y, y', d) \sim D^{\text{pref}}} \left[ \max(0, 1 - d \cdot (r_\theta(y) - r_\theta(y'))) \right].
\end{equation}

Once the reward model is trained, it is used to update the policy parameters $\phi$ via Proximal Policy Optimization (PPO). In order to maintain the capabilities obtained from the pre-training and instruct tuning stages, corresponding loss terms are incorporated into the objective, resulting in an expanded objective function. $\phi$ is trained by maximizing the following objective:

\begin{equation}
\begin{aligned}
J^{\mathrm{RL}}(\phi)= & \mathbb{E}_{(x, y) \sim D^{\mathrm{RL}}}\left[r_\theta(x, y)-\lambda_{\mathrm{IT}} \log \left(\pi_\phi^{\mathrm{RL}}(y \mid x) / \pi_\phi^{\text{inst}}(y \mid x)\right)\right]+ \\
& \lambda_{\mathrm{PT}} \mathbb{E}_{x \sim D^{\text {pretrain }}}\left[\log \left(\pi_\phi^{\mathrm{RL}}(x)\right)\right],
\end{aligned}
\end{equation}

where $r_\theta(x, y)$ is the reward for a generated response $y$ given a prompt $x$ under the reinforcement learning updated policy $\pi_\phi^{\mathrm{RL}}$, $\lambda_{\mathrm{IT}}$ and $\lambda_{\mathrm{PT}}$ are regularization terms that keeps the RL policy close to the instruct tuning and pre-training distribution. This hybridized reinforcement learning approach enables \textsc{BatGPT} to leverage the nuanced understanding of humans and the robust consistency of AI systems. As such, \textsc{BatGPT} can generate more beneficial, better-aligned, and safer outputs, making it suitable for a wide range of applications.

\subsection{Parameter Expansion}

In the development of \textsc{BatGPT}, a crucial step involves the effective expansion of parameters, allowing the newly initialized model to retain the same functional characteristics as its source. This section elaborates on the expansion strategy employed to achieve this feat, focusing on the two key stages of width and depth expansion. 

\paragraph{Function Preserving Initialization (FPI)}

For width expansion, we start with a source weight matrix $\mathbf{Q} \in \mathbb{R}^{d_{\mathrm{in}}^q * d_{\mathrm{out}}^q}$ that has been trained to provide a set of functional parameters. The goal of width expansion is to form a new target weight matrix $\mathbf{V} \in \mathbb{R}^{d_{\mathrm{in}}^v * d_{\mathrm{out}}^v}( d_{\mathrm{in}}^v> d_{\mathrm{in}}^q , d_{\mathrm{out}}^v> d_{\mathrm{out}}^q)$. With an aim to ensure functional consistency between the source and target models, \textsc{BatGPT} adopts Function Preserving Initialization (FPI). This technique, as introduced in \cite{chen2015net2net} and \cite{chen-etal-2022-bert2bert}, guarantees that the initialized model produces the same output for the same input as the original model, achieved through two critical mapping functions, $m_{\text{in}}$ and $m_{\text{out}}$. The mapping functions are defined as follows:
\begin{equation}
\begin{aligned}
m_{\text{in}}(i) & = \begin{cases}i & i \in\left[1, d_{\text{in}}^q\right] \\
s\left(\left\{1,2, \ldots, d_{\text{in}}^q\right\}\right) & i \in\left(d_{\text{in}}^q, d_{\text{in}}^v\right]\end{cases} \\
m_{\text{out}}(j) & = \begin{cases}j & j \in\left[1, d_{\text{out}}^q\right] \\
s\left(\left\{1,2, \ldots, d_{\text{out}}^q\right\}\right) & j \in\left(d_{\text{out}}^q, d_{\text{out}}^v\right]\end{cases}.
\end{aligned}
\end{equation}

These functions map the i-th column and the j-th row of the target weight matrix \( \mathbf{V} \) to corresponding parameters of the source weight matrix \( \mathbf{Q} \), utilizing uniform sampling denoted by \( s(\cdot) \). The actual expansion of weights is represented by the expression:
\begin{equation}
\mathbf{V} = \text{EX}(\mathbf{Q}; m_{\text{in}}, m_{\text{out}}).
\end{equation}
This process encompasses two defined stages for expanding in-dimension and out-dimension, respectively.
\begin{equation}
\begin{gathered}
M_{m_{\text{in}}(i)}=\sum_{i^{\prime}=1}^{d_{\text{in}}^v} \mathbb{I}\left(m_{\text{in}}\left(i^{\prime}\right)=m_{\text{in}}(i)\right) \\
\widetilde{\mathbf{V}}_{(i, *)}=\frac{1}{M_{m_{\text{in}}(i)}} \mathbf{Q}_{\left(m_{\text{in}}(i), *\right)}, \\
\mathbf{V}_{(*, j)}=\widetilde{\mathbf{V}}_{\left(*, m_{\text{out}}(j)\right)},
\end{gathered}
\end{equation}

Here, \( \mathbb{I}(\cdot) \) functions as an indicator, and \( M_{m_{\text{in}}(i)} \) re-scales the original parameters, ensuring the function-preserving attribute.

\paragraph{Progressive Stacking}

For depth expansion, \textsc{BatGPT} employs a \textit{progressive stacking} method from \cite{gong2019efficient}. This concept leverages the principle that training shallow models is typically faster than training deep ones. Thus, the overall training loop of deep models can be accelerated by iteratively stacking them from shallower structures.  Specifically, given a source model $M_0$ with initial depth $L/2^k$, the goal of depth expansion is to progressively increase the number of layers until reaching the desired depth $L$, while optimizing training efficiency. The iteration loop of \textit{progressive stacking} can be summarized as follows:

\begin{algorithm}
\caption{Progressive stacking for depth expansion}
\label{alg:progressive_stacking}
\begin{algorithmic}
\State Initialization: \(M_0' \leftarrow\) InitModel(\(L/2^k\))
\State \(M_0 \leftarrow\) Train(\(M_0\)) \textit{// Train from scratch.}
\For{\(i \leftarrow 1\) \textbf{to} \(k\)}
    \State \(M_i' \leftarrow\) StackLayers(\(M_{i-1}\)) \textit{// Duplicate and stack the layers.}
    \State \(M_i \leftarrow\) Train(\(M_i'\)) \textit{// \(M_i\) has \(L/2^{k-i}\) layers.}
\EndFor
\State \textbf{return} \(M_k\)
\end{algorithmic}
\end{algorithm}

In essence, the progressive stacking approach begins with a shallow model, and then successively doubles its depth through a series of training and stacking iterations. This method has shown significant efficacy in reducing training time without sacrificing the quality of the final deep model.

In the context of \textsc{BatGPT}, the parameter expansion techniques played an essential role in the development of various model sizes. Specifically, the 15B version of \textsc{BatGPT} was trained from scratch using a 1-trillion-token English and Chinese bilingual corpus, serving as the foundational model. Subsequent larger models, including the 30B and 70B versions, were initialized from this 15B base. By jointly utilizing width and depth expansion, these more complex models were efficiently constructed, leveraging the already-trained parameters. This approach not only expedited the training process but also ensured a level of consistency and stability across different model scales.

\section{Performance}

\subsection{Evaluation on CMMLU}

The CMMLU \cite{li2023cmmlu} is a comprehensive benchmark specifically designed to assess language model capability over a variety of Chinese-related subjects. This benchmark evaluates models across diverse categories, including humanities, social sciences, STEM (science, technology, engineering, and mathematics), and other areas. The evaluation results reported in CMMLU is presented in Table~\ref{table:CMMLU}. \textsc{BatGPT} ranked second among all Chinese-oriented language models, exhibiting promising performance. \textsc{BatGPT}-15B achieved the highest average accuracy in both the STEM and Others categories with scores of 33.49\% and 42.14\% respectively. The model demonstrated consistent high performance across the other categories as well: it scored 35.38\% in Humanities, 36.31\% in Social Science, and 37.00\% in China-specific topics.

\begin{table}[t]
\centering
\small
\caption{Performance comparison of \textsc{BatGPT} and other Chinese-oriented large language models (LLMs) on the CMMLU benchmark. Results are presented as average accuracy within each categories based on a five-shot experiment setting.}
\label{table:CMMLU}
\begin{tabular}{lcccccc}
\toprule
Model             & STEM           & Humanities & Social Science & Other & China-specific & \textbf{Average}  \\ \midrule
MOSS-SFT-16B      & 27.23          & 30.41      & 28.84          & 32.56 & 28.68          & 29.57   \\ 
\textsc{BatGPT}-15B        & \textbf{33.49} & 35.38      & 36.31          & \textbf{42.14} & 37.00          & 36.72   \\ 
Chinese-LLaMA-13B & 27.12          & 33.18      & 34.87          & 35.10 & 32.97          & 32.63   \\ 
Chinese-GLM-10B   & 25.49          & 27.05      & 27.42          & 29.21 & 28.05          & 27.26   \\ 
Chinese-LLaMA-7B  & 25.79          & 27.45      & 26.35          & 26.06 & 25.45          & 26.36   \\ 
ChatGLM-6B        & 32.35          & 39.22      & 39.65          & 38.62 & 37.70          & 37.48   \\ \midrule
Random            & 25.00          & 25.00      & 25.00          & 25.00 & 25.00          & 25.00   \\
\bottomrule
\end{tabular}
\end{table}

Additional results from the CMMLU benchmark, shown in Table~\ref{table:CMMLU}, provide an insightful evaluation of \textsc{BatGPT}-15B's zero-shot capabilities. When prompted with a direct answer (DA), the model achieves an accuracy of 33.74\% on the STEM subset and 38.48\% overall. In contrast, under the chain-of-thought (COT) prompt, the accuracies are 34.47\% and 35.91\% respectively. The fact that the COT prompts does not consistently outperform the DA ones might be attributed to the scale of \textsc{BatGPT}-15B, suggesting that a larger version might harness the potential of COT more effectively.

Overall, \textsc{BatGPT}'s robust performance across various topics and its ability to handle different types of prompts demonstrate its wide-ranging capabilities and its suitability for a broad spectrum of applications.

\subsection{Evaluation on \textsc{C-Eval}}

\begin{table}[h]
\centering
\begin{minipage}{.45\textwidth}
\centering
\caption{Zero-shot accuracy on CMMLU STEM subset, and full set, with direct answer (DA) prompt and chainof-thought (COT) prompt.}
\label{table:CMMLU}
\begin{tabular}{lcc}
\toprule
Category & DA       & COT      \\
\midrule
STEM     & 33.74\%  & 34.47\%  \\
Overall  & 38.48\%  & 35.91\%  \\
\bottomrule
\end{tabular}
\end{minipage}%
\hfill
\begin{minipage}{.5\textwidth}
\centering
\caption{Evaluation results of \textsc{BatGPT} on \textsc{C-Eval}. \textbf{Avg(Hard)} denotes the averaged accuracy over the \textsc{C-Eval Hard} subset.}
\label{table:CEval}
\begin{tabular}{lc}
\toprule
Category               & Accuracy           \\
\midrule
STEM                   & 50.5\%             \\
Social Science         & 72.1\%             \\
Humanities             & 60.7\%             \\
Others                 & 53.3\%             \\
\textbf{Average}       & 57.4\%             \\
\textbf{Avg(Hard)}     & 36.9\%             \\
\bottomrule
\end{tabular}
\end{minipage}
\end{table}

\textsc{C-Eval} \cite{huang2023ceval} is another comprehensive Chinese LLM evaluation suite developed to critically assess the capabilities of LLMs, particularly in a Chinese context. Comprising multiple-choice questions across four difficulty levels and spanning 52 diverse disciplines, it serves as a nuanced measure of model proficiency in various subjects. The evaluation results of \textsc{BatGPT}-15B on C-Eval are presented in Table~\ref{table:CEval}. In this assessment, \textsc{BatGPT} achieved an average accuracy of 57.4\%, with particularly strong performance in the Social Science category at 72.1\%. \textsc{BatGPT} also demonstrated consistent proficiency across various domains, with scores of 60.7\% in Humanities, 50.5\% in STEM, and 53.3\% in the Others category. Of particular interest is the model's performance on the \textsc{C-Eval Hard} subset — a collection of exceptionally challenging subjects within the suite designed to assess non-trivial reasoning abilities. In this stringent evaluation, \textsc{BatGPT}-15B secured an accuracy of 36.9\%, underscoring its potential in tackling intricate and nuanced tasks. Collectively, these results highlight the robust generalization ability of \textsc{BatGPT}, confirming its adaptability and effectiveness across a diverse range of Chinese-centric topics and difficulty levels.


\section{Conclusion}

In this paper, we introduce \textsc{BatGPT}, a large-scale language model jointly developed by Wuhan University and Shanghai Jiao Tong University. By leveraging a bidirectional autoregressive architecture, \textsc{BatGPT} addresses the limitations of existing models, such as limited memory and hallucinations, allowing it to capture complex dependencies more efficiently.
The training of \textsc{BatGPT} incorporates an effective parameter expansion method that leverages pre-training of smaller models and reinforcement learning from both AI and human feedback. This approach enhances the model's alignment performance and enables it to generate text that is highly fluent, coherent, and contextually relevant.
However, it is important to acknowledge that challenges and limitations still exist. Ongoing research and development efforts are essential to further refine and improve the capabilities of language models like \textsc{BatGPT}. Ethical considerations regarding bias, fairness, and responsible use also need to be addressed to ensure the deployment of these models benefits society as a whole.
As research and development continue, we can expect further breakthroughs in language understanding and generation, paving the way for even more sophisticated and impactful language models in the future.

\bibliography{custom}

\end{document}